\documentclass{llncs}
\usepackage{graphicx}
\usepackage{multicol}
\usepackage{amsmath, amssymb, amscd, tikz}
\usepackage{makeidx}

\newtheorem{Def}{Definition}

\begin{document}
%
%
\title{Fuzzy Object-Oriented Dynamic Networks. II}
\titlerunning{Fuzzy Object-Oriented Dynamic Networks. II}  
%
\author{D. A. Terletskyi\inst{1}, A. I. Provotar\inst{2}}
\authorrunning{} 
%
\tocauthor{}
\institute{Taras Shevchenko National University of Kyiv, Kyiv, 03680, Ukraine
\email{dmytro.terletskyi@gmail.com},\\
\and
University of Rzeszow, Rzeszow, 35--310, Poland\\
\email{aprowata1@bigmir.net}}

\maketitle              

\begin{abstract}
This article generalizes object-oriented dynamic networks to the fuzzy case, which allows one to represent knowledge on objects and classes of objects that are fuzzy by nature and also to model their changes in time. Within the framework of the approach described, a mechanism is proposed that makes it possible to acquire new knowledge on the basis of basic knowledge and considerably differs from well-known methods used in existing models of knowledge representation. The approach is illustrated by an example of construction of a concrete fuzzy object-oriented dynamic network.

\keywords{ fuzzy object, class of fuzzy objects, modifier, exploiter.}
\end{abstract}
\section*{Introduction}

At present, one of important problems in the field of knowledge representation is the generalization of knowledge representation models (KRMs) to the fuzzy case using a powerful tool such as fuzzy sets \cite{Zadeh}. The need for this generalization is stipulated by the fact that the body of knowledge available to man is inexact, incomplete, or diffused in a sense \cite{Leung-Wong}, \cite{Berzal}. Since fuzzy sets form an efficient apparatus for modeling objects and processes of such a nature, it is expedient to generalize KRMs and use them. There are many similar generalizations, in particular, fuzzy logic, fuzzy semantic networks, fuzzy production models, fuzzy neural networks, fuzzy ontologies, fuzzy frames, fuzzy UML, etc. First of all, these KRMs are theoretical models that can practically be implemented within the framework of concrete intelligent information systems (IISs) with applying some programming paradigm or other. In recent years, object-oriented programming (OOP) is most popular and widely used. The majority of well-known KRMs can be implemented in concrete IISs with the help of OOP. Moreover, KRMs such as frames or scripts are ideologically close to OOP. This approach is also rather efficient in constructing and controlling databases \cite{Berzal}, \cite{Marin}. However, despite all its advantages, the possibilities of representation of objects that are fuzzy by nature are not provided in it. Therefore, the improvement of this approach with the help of fuzzy sets is a topical problem \cite{Leung-Wong}, \cite{Berzal}, \cite{Ndousse}.

To represent fuzzy and diffused knowledge, this article proposes to generalize such KRMs to object-oriented dynamic networks \cite{Terletskyi-1}. To construct fuzzy object-oriented dynamic networks, the concepts of a fuzzy object, a class of fuzzy objects, and also operations over them \cite{Terletskyi-2} are used.

\section*{Object-Oriented Representation of Fuzzy Knowledge}

Let us consider some features of the object-oriented approach to fuzzy knowledge representation whose basic concepts are fuzzy objects, classes of fuzzy objects, and relations between them.

Each real or abstract object has some characteristic properties that can be represented in the form of attributes of its object model. An object is called fuzzy if it has at least one fuzzy property, i.e., a property expressed with the help of a fuzzy set \cite{Ndousse}, \cite{Ma-Zhang}, \cite{Zhang-Ma}.

Since there are objects of identical nature, the concept of a class is introduced within the framework of which they are defined. Thus, each object is a concrete example of some class (type) of objects. In OOP, a class is considered as a certain prototype or an abstract object for constructing concrete objects of some type. A class of objects is called fuzzy if it has at least one fuzzy property \cite{Ma-Zhang}, \cite{Zhang-Ma}.

A class of objects can be considered from two different positions, namely, extensional and intensional \cite{Ma-Zhang}, \cite{Zhang-Ma}, \cite{Bordogna}. In the first case, it is specified with the help of a definite number of objects that belong to it and, in the second case, it is specified with the help of enumeration of its own attributes and their possible values. It may appear that, despite their distinctions, the both methods of representation of a class of objects are finally reduced to same result, namely, to a definite class of objects. In fact, the classes of objects obtained with the help of these approaches can have fundamental differences.

If the extensional approach is used, then the obtained class of objects will depend on the objects themselves on the basis of which it is defined. If all attributes of all objects are defined, i.e., each property of each object has a concrete value or a range of values, then the class defined on the basis of these objects will also be defined. Otherwise, it will be partially defined or undefined.

If the intensional approach is applied, then the definiteness of the obtained class of objects will depend on the definition of its attributes, i.e., if all the properties of a class of objects are completely defined, then, as is obvious, the class will also be defined and, otherwise, it will by partially defined or undefined.

It is also important to know the author and purpose of definition of some class of objects or other. Assume that there are three squares of different sizes. If a certain class of squares is defined on their basis, then we obtain the class describing squares of only these three sizes. If we assume that a class of squares should be described without alluding to any concrete squares, then, theoretically, a class of squares can be defined that describes, for example, the squares from the previous case. If concrete squares are not considered, then, most likely, a class will be defined that described squares of all possible sizes. Proceeding from the last assertion, we can draw the conclusion that both methods of representation of classes of objects are useful but the mentioned nuances should be taken into account in using them.

In OOP, objects and classes are related with the help of well-defined relations within the framework of the corresponding hierarchy constructed with the help of an inheritance mechanism \cite{Stroustrup}. In the case of fuzzy objects and classes of fuzzy objects, the membership of an object in a class can be fuzzy (partial), i.e., with some measure. The following relations between fuzzy objects and their classes are singled out: generalization, aggregation, and association \cite{Ndousse}, \cite{Zhang-Ma}. The generalization relation shows that some class is a subclass or an example of another class. The aggregation relation shows that some class is a component part of another class. The association relation reflects certain semantic relations between classes that are not related among themselves by a generalization or aggregation relation.

Within the framework of the object-oriented approach to the representation of fuzzy knowledge, we will consider an KRM underlain by frames or, more exactly, its generalization to the fuzzy case \cite{Graham-1}, \cite{Graham-2}. Frames are similar to OOP in many respects, but, in this case, the emphasis in frame systems is on the infrastructure of their object domains constructed from objects, classes, and relations between them, whereas, in OOP, the attention is accented on the messaging between concrete objects \cite{Brachman-Levesque}. As well as classes and objects in the object-oriented approach to knowledge representation, frames are also commonly called fuzzy if they have at least one fuzzy attribute or partially inherit other frames \cite{Graham-1}. The introduction of the conception of partial inheritance leads to considerable changes in this mechanism as a result of addition of new possibilities to it.

\section*{Fuzzy Object-Oriented Dynamic Networks}

To construct fuzzy object-oriented dynamic networks, we will use the conception of object-oriented dynamic networks, which is described in \cite{Terletskyi-1}, and also generalizations of all its component parts to the fuzzy case \cite{Terletskyi-2}.
\begin{Def}
A fuzzy object-oriented dynamic network FOODN is an object-oriented dynamic network for which at least one of the following conditions is fulfilled:
\begin{enumerate}
\item $\exists A_k,\dots,A_m\in Î=\{A_1,\dots,A_n\}$, where $1\leq k\leq m\leq n$ and $A_k,\dots,A_m$ are fuzzy objects.
\item $\exists T_p,\dots,T_q\in C=\{T_1,\dots,T_w\}$, where $1\leq p\leq q\leq w$ and $T_p,\dots,T_q$ are classes of fuzzy objects;
\item $\exists R_i,\dots,R_j\in R=\{R_1,\dots,R_v\}$, where $1\leq i\leq j\leq v$ and $R_i,\dots,R_j$ are fuzzy relations between fuzzy objects and classes of fuzzy objects.
\end{enumerate}
\end{Def}

As relations between fuzzy objects and classes of fuzzy objects that form the set $R$, we consider the three above-mentioned types of relations used in the object-oriented approach to the representation of fuzzy knowledge. Note that, to some extent, these types of relations are inherent in semantic networks, frames, and scripts and are described in detail in the corresponding literature, in particular, in \cite{Brachman-Levesque}, \cite{Negnevitsky}.

To illustrate FOODN, we will use the example of OODN presented in \cite{Terletskyi-1} since it is intuitive and rather easily understandable and also reflects basic distinctive features of the proposed knowledge representation model.

{\bf Example 1.} Let us consider the classes $T(Pg)$, $T(Rb)$ and $T(Sq)$ of fuzzy objects that describe the class of fuzzy convex polygons, class of fuzzy rhombuses, and class of fuzzy squares, respectively. We define their specifications and signatures as follows:
\begin{align*}
T(Pg)=&(P(Pg),F(Pg))=(p_1(Pg),\dots,p_4(Pg),f_1(Pg)),\\
T(Rb)=&(P(Rb),F(Rb))=(p_1(Rb),\dots,p_6(Rb),f_1(Rb),f_2(Rb)),\\
T(Sq)=&(P(Sq),F(Sq))=(p_1(Sq),\dots,p_6(Sq),f_1(Sq),f_2(Sq)).
\end{align*}
We present the specifications and signatures of the classes of fuzzy objects $T(Pg)$, $T(Rb)$ and $T(Sq)$ with the help of Tables~\ref{tab-1} and \ref{tab-2}, respectively.
\begin{table}[!h]
\centering{
\caption{Specification of the Classes of Fuzzy Objects $T(Pg)$, $T(Rb)$ and $T(Sq)$}
\label{tab-1}
\begin{tabular}{p{2cm}p{3.8cm}p{2.1cm}p{2.1cm}p{2cm}}
\hline\noalign{\smallskip}
\multicolumn{1}{c}{Property $p_i$} & Semantic Values of Properties of Classes of Objects & \multicolumn{3}{c}{Values of Properties of Classes of Objects}\\
 & & \multicolumn{1}{c}{$T(Pg)$} & \multicolumn{1}{c}{$T(Rb)$} & \multicolumn{1}{c}{$T(Sq)$}\\
\noalign{\smallskip}
\hline
\noalign{\smallskip}
\multicolumn{1}{c}{$p_1$} & Number of sides & \multicolumn{1}{c}{$4$} & \multicolumn{1}{c}{$4$} & \multicolumn{1}{c}{$4$}\\
\multicolumn{1}{c}{$p_2$} & Sizes of sides & \multicolumn{1}{c}{fuzzy} & \multicolumn{1}{c}{fuzzy} & \multicolumn{1}{c}{fuzzy}\\
\multicolumn{1}{c}{$p_3$} & Number of angles & \multicolumn{1}{c}{$4$} & \multicolumn{1}{c}{$4$} & \multicolumn{1}{c}{$4$}\\
\multicolumn{1}{c}{$p_4$} & Grade measures of angles & \multicolumn{1}{c}{$(0^\circ,180^\circ)$} & \multicolumn{1}{c}{$(0^\circ,180^\circ)$} & \multicolumn{1}{c}{$90^\circ,90^\circ,90^\circ,90^\circ$}\\
\multicolumn{1}{c}{$p_5$} & Equality of all sides & \multicolumn{1}{c}{--} & \multicolumn{1}{c}{$1$} & \multicolumn{1}{c}{$1$}\\
\multicolumn{1}{c}{$p_6$} & Equality of all angles & \multicolumn{1}{c}{--} & \multicolumn{1}{c}{fuzzy} & \multicolumn{1}{c}{$1$}\\
\hline
\end{tabular}}
\end{table}
\begin{table}[!h]
\centering{
\caption{Signature of the Classes of Fuzzy Objects $T(Pg)$, $T(Rb)$ and $T(Sq)$}
\label{tab-2}
\begin{tabular}{p{1.8cm}p{3cm}p{3.2cm}p{1.8cm}p{1.7cm}}
\hline\noalign{\smallskip}
\multicolumn{1}{c}{Method $f_i$} & Calculus Function & \multicolumn{3}{c}{Body of the Method of Classes of Objects}\\
 & & \multicolumn{1}{c}{$T(Pg)$} & \multicolumn{1}{c}{$T(Rb)$} & \multicolumn{1}{c}{$T(Sq)$}\\
\noalign{\smallskip}
\hline
\noalign{\smallskip}
\multicolumn{1}{c}{$f_1$} & Perimeter of a figure & \multicolumn{1}{c}{$f_1(Pg)=\sum_{i=1}^n a_i$} & \multicolumn{1}{c}{$f_1(Rb)=4a$} & \multicolumn{1}{c}{$f_1(Sq)=4a$}\\
\multicolumn{1}{c}{$f_2$} & Area of a figure & \multicolumn{1}{c}{--} & \multicolumn{1}{c}{$f_2(Rb)=a^2\sin\alpha$} & \multicolumn{1}{c}{$f_2(Sq)=a^2$}\\
\hline
\end{tabular}}
\end{table}

Analyzing Table \ref{tab-1}, note that the value of the property $p_2$ is specified as fuzzy, i.e., sizes of sides of figures are represented in the form of fuzzy sets. The definition of the classes $T(Pg)$, $T(Rb)$ and $T(Sq)$ is intensional since, in this situation, there is no need to consider the classes describing concrete types of polygons, rhombuses, and squares. For the same reason, for the properties $p_4(Pg) $and $p_4(Rb)$ concrete values are not specified and only ranges of possible values are given.

Let us consider concrete objects of the classes $T(Rb)$ and $T(Sq)$ namely, the fuzzy rhombus $Rb_1$ and fuzzy square $Sq_1$. We define the values of their properties and methods taking into account the specifications and signatures of their classes with the help of Tables~\ref{tab-3} and \ref{tab-4} respectively.
\begin{table}
\centering{
\caption{Specifications of the Fuzzy Objects $Rb_1$ and $Sq_1$}
\label{tab-3}
\begin{tabular}{cp{4.7cm}p{4.8cm}}
\hline\noalign{\smallskip}
\multicolumn{1}{c}{Property $p_i$} & \multicolumn{2}{c}{Values of Properties of Classes of Objects}\\
& \multicolumn{1}{c}{Rhombus $Rb_1$} & \multicolumn{1}{c}{Square $Sq_1$}\\
\noalign{\smallskip}
\hline
\noalign{\smallskip}
\multicolumn{1}{c}{$p_1$} & \multicolumn{1}{c}{$4$} & \multicolumn{1}{c}{$4$}\\
\multicolumn{1}{c}{$p_2$} & $(\{1.8/0.9+2/1+2.1/0.95\},cm)$, $(\{1.8/0.9+2/1+2.1/0.95\},cm)$, $(\{1.8/0.9+2/1+2.1/0.95\},cm)$, $(\{1.8/0.9+2/1+2.1/0.95\},cm)$ & $(\{2.7/0.85+3/1+3.1/0.95\},cm)$, $(\{2.7/0.85+3/1+3.1/0.95\},cm)$, $(\{2.7/0.85+3/1+3.1/0.95\},cm)$, $(\{2.7/0.85+3/1+3.1/0.95\},cm)$\\
\multicolumn{1}{c}{$p_3$} & \multicolumn{1}{c}{$4$} & \multicolumn{1}{c}{$4$}\\
\multicolumn{1}{c}{$p_4$} & \multicolumn{1}{c}{$95^\circ,85^\circ,95^\circ,85^\circ$} & \multicolumn{1}{c}{$90^\circ,90^\circ,90^\circ,90^\circ$}\\
\multicolumn{1}{c}{$p_5$} & \multicolumn{1}{c}{$1$} & \multicolumn{1}{c}{$1$}\\
\multicolumn{1}{c}{$p_6$} & \multicolumn{1}{c}{$0.8$} & \multicolumn{1}{c}{$1$}\\
\hline
\end{tabular}}
\end{table}
\begin{table}
\centering{
\caption{Signatures of the Fuzzy Objects $Rb_1$ and $Sq_1$}
\label{tab-4}
\begin{tabular}{cp{6cm}p{6cm}}
\hline\noalign{\smallskip}
\multicolumn{1}{c}{Method $f_i$} & \multicolumn{2}{c}{Body of the Method of Classes of Objects}\\
& \multicolumn{1}{c}{Rhombus $Rb_1$} & \multicolumn{1}{c}{Square $Sq_1$}\\
\noalign{\smallskip}
\hline
\noalign{\smallskip}
\multicolumn{1}{c}{$f_1$} & \multicolumn{1}{c}{$f_1(Rb_1)=4a$} & \multicolumn{1}{c}{$f_1(Sq_1)=4a$}\\
\multicolumn{1}{c}{$f_2$} & \multicolumn{1}{c}{$f_2(Rb_1)=a^2\sin\alpha$} & \multicolumn{1}{c}{$f_2(Sq_1)=a^2$}\\
\hline
\end{tabular}}
\end{table}

We now pass to the direct construction of a fuzzy object-oriented dynamic network for the fuzzy objects $Rb_1$ and $Sq_1$ and the classes of fuzzy objects $T(Pg)$, $T(Rb)$ and $T(Sq)$. It is obvious that the set of fuzzy objects is $O=\{Rb_1,Sq_1\}$, and the set of classes of fuzzy objects is $C=\{T(Pg),T(Rb),T(Sq)\}$. The classes of fuzzy objects $T(Rb)$ and $T(Sq)$ are examples of the class of fuzzy objects $P(Pg)$, and any square is a rhombus. According to these conclusions, the set of relations $R$ is of the form
\begin{gather*}
R=\left\{Rb_1\xrightarrow{instance-of}T(Rb),\ Sq_1\xrightarrow{instance-of}T(Sq), \right.\\
\left.T(Rb)\xrightarrow{a-kind-of}T(Pg),\ T(Sq)\xrightarrow{a-kind-of}T(Pg),\ T(Sq)\xrightarrow{is-a}T(Rb)\right\},
\end{gather*}
which can also be written as
\begin{gather*}
R=\{Rb_1\in T(Rb),\ Sq_1\in T(Sq),\ T(Pg)\subseteq T(Rb),\ T(Pg)\subseteq T(Sq), \\
T(Rb)\subseteq T(Sq)\}.
\end{gather*}

We will define a set of exploiters $E$ using the universal operations over objects and classes that are proposed in \cite{Terletskyi-3}; their practical application to fuzzy objects and classes of fuzzy objects is considered in \cite{Terletskyi-2}. Thus, the set of exploiters $E$ consists of the operations of union, intersection, difference, symmetric difference, and also the operation of cloning fuzzy objects and classes of fuzzy objects, i.e.,
\[E=\left\{E_1^n,E_2^n,E_2^3,E_4^2,E_5^1\right\}=\left\{\cup,\ \cap,\ \setminus,\ \div,\ Clone_i\right\},\]
where $E_1^n$ is the union operation, $E_2^n$ is the intersection operation, $E_3^2$ is the difference operation, $E_4^2$ is the operation of symmetric difference, and $E_5^1$ is the cloning operation. The inferior index of each exploiter is its number in the signature, and the upper index is its arity. These operations are partly universal since they can be applied to all fuzzy objects and their classes.

We define a set of modifiers $M$ as follows:
\begin{gather*}
M=\left\{M_1(T(Sq)):T(Sq)\rightarrow T(Rb),\ M_1(T(Rb)):T(Rb)\rightarrow T(L_1), \right.\\
M_2(T(Rb)):(T(Rb))\rightarrow T(Sq),\ M_1(T(Pg)):T(Pg)\rightarrow T(L),\\
\left.M_1(Sq_1):Sq_1\rightarrow Rb_1,\ M_1(Rb_1):Rb_1\rightarrow L_{1_1},\ M_2(Rb_1):Rb_1\rightarrow Sq_1\right\},
\end{gather*}
where
\[M_1(T(Sq))=M_1(P(Sq))=M_1(p_6(S1))=M_1(1)=0.8\]
is the partial modifier that changes the property $p_6(Sq)$ of the class of fuzzy squares $T(Sq)$ and, thereby, transforms it into the class of fuzzy rhombuses $T(Rb)$;
\[M_1(T(Rb))=M_1(p_1(Rb))=M_1(4,sides)=(3,of\ segment)\]
is the partial modifier that changes the property $p_1(Rb)$ of the class of fuzzy rhombuses $T(Rb)$ by transforming it into some class of fuzzy broken lines $T(L_1)$;
\[M_2(T(Rb))=M_2(P(Rb))=M_2(p_6(Rb))=M_2(0.8)=1\]
is the partial modifier that changes the property $p_6(Rb)$ of the class of fuzzy rhombuses $T(Rb)$ by transforming it into the class of fuzzy squares $T(Sq)$;
\[M_1(T(Pg))=M_1(p_1(Pg))=M_1(4,sides)=(3,of\ segment)\]
is the partial modifier that changes the property $p_1(Pg)$ of the class of fuzzy convex polygons $T(Pg)$ by transforming it into some class of fuzzy broken lines $T(L)$;
\begin{gather*}
M_1(Sq_1)=M_1(p_4(Sq_1),p_6(Sq_1))=M_1((90^\circ,90^\circ,90^\circ,90^\circ),1)=\\
=((95^\circ,85^\circ,95^\circ,85^\circ),0.8)
\end{gather*}
is the partial modifier that changes the properties $p_4(Sq_1)$ and $p_6(Sq_1)$ by transforming the fuzzy square $Sq_1$ into the fuzzy rhombus $Rb_1$;
\[M_1(Rb_1)=M_1(p_1(Rb_1))=M_1(4,sides)=(3,of\ segment)\]
is the partial modifier that changes the property $p_1(Rb_1)$ of the fuzzy rhombus $Rb_1$ by transforming it into the fuzzy broken line $L_{1_1}$;
\begin{gather*}
M_2(Rb_1)=M_2(p_4(Rb_1),p_6(Rb_1))=M_2((95^\circ,85^\circ,95^\circ,85^\circ),0.8)=\\
=((90^\circ,90^\circ,90^\circ,90^\circ),1)
\end{gather*}
is the partial modifier that changes the properties $p_4(Rb_1)$ and $p_6(Rb_1)$ of the fuzzy rhombus $Rb_1$ by transforming it into the fuzzy square $Sq_1$.

Analyzing the modifiers from the set $M$, note that some of them simultaneously modify several properties of fuzzy objects and classes of fuzzy objects. The properties being modified are non-randomly chosen since all of them are related in one way or another, and a change in one of them must imply corresponding changes in the others. This follows from the reflection principle considered in \cite{Terletskyi-2}.

\begin{figure}[h]
\centering
\includegraphics[width=0.95\textwidth]{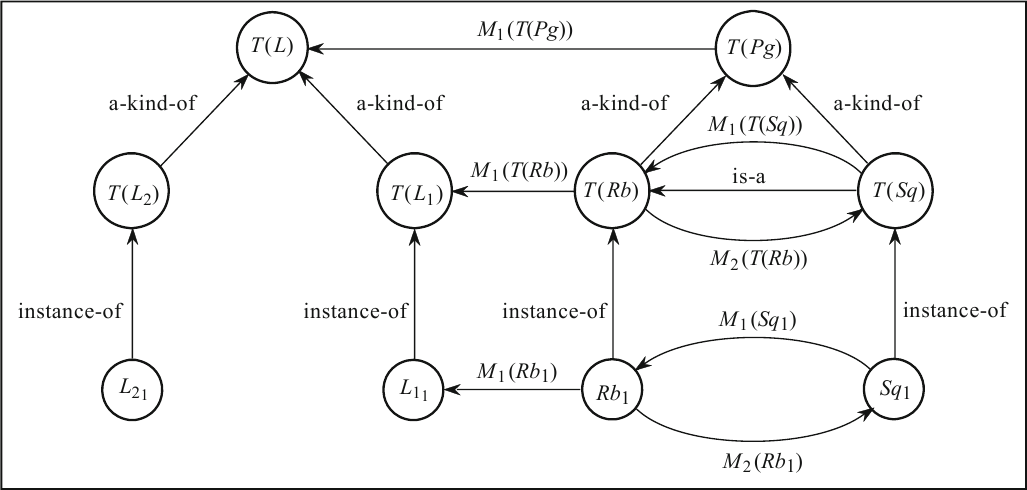}
\caption{The FOODN fragment for the fuzzy objects $Rb_1$ and $Sq_1$ and classes of fuzzy objects $T(Rb)$, $T(Sq)$ and $T(Pg)$.}
\label{fig-1}
\end{figure}

For simplicity and also for the understanding of the structure and essence of the fuzzy object-oriented dynamic network constructed for the objects $Rb_1$ and $Sq_1$ and classes of objects $T(Rb)$, $T(Sq)$ and $T(Pg)$ we represent it in the form of a connected digraph (Fig.~\ref{fig-1}).

The graph vertices are considered to be the fuzzy objects $Rb_1$, $Sq_1$, $L_{1_1}$, and $L_{2_1}$ and the classes of fuzzy objects $T(Rb)$, $T(Sq)$, $T(Pg)$, $T(L)$, $T(L_1)$ and $T(L_2)$ and relationships between them, i.e., the relations \emph{a-kind-of}, \emph{instance-of}, and \emph{is-a}, and the modifications $M_1(T(Sq))$, $M_1(T(Rb))$, $M_2(T(Rb))$, $M_1(T(Pg))$, $M_1(Rb_1)$, $M_2(Rb_1)$ and $M_1(Sq_1)$. The relations partly connect objects and classes of objects among themselves and form some hierarchy. This part of the network is static since only the structure of knowledge on some objects and classes of objects can be represented with its help.

Modifications can be considered as some new type of relations between objects and classes of objects that is presented as \emph{modification-of}. Proceeding from Fig.~\ref{fig-1}, note that
\[T(Sq)\xrightarrow{modification-of}T(Rb)\]
and, vice versa,
\[T(Rb)\xrightarrow{modification-of}T(Sq).\]
But, in contrast to the other types of relations, modifications are not completely static since modifiers are some methods that can be applied to objects or classes of objects, and, as a result, they change them.

If we consider modifications of some fuzzy square $Sq_k$ with the help of some modifier $M_1(Sq_k)$ that transforms it into some fuzzy rhombus $Rb_m$, then, as a result, from a fuzzy square, we obtain a fuzzy rhombus rather than both these figures simultaneously as is illustrated in Fig.~\ref{fig-1} by the example of modification of the fuzzy square $Sq_1$. The FOODN presented in Fig.~\ref{fig-1} is constructed on the basis of the objects $Sq_1$ and $Rb_1$ and, hence, they are component parts of the network. This method of graphical representation of the process of modification is chosen with a view to showing changes in the essence of an object or a class of objects after applying a modifier to it. It may appear that this solution is not most optimal, but if a modification of an object is considered as a process of creating a new class of objects and not just as a change in this concrete object, then the chosen method of graphical representation of the process of modification is sufficiently substantiated.

\begin{figure}[h]
\centering
\includegraphics[width=0.95\textwidth]{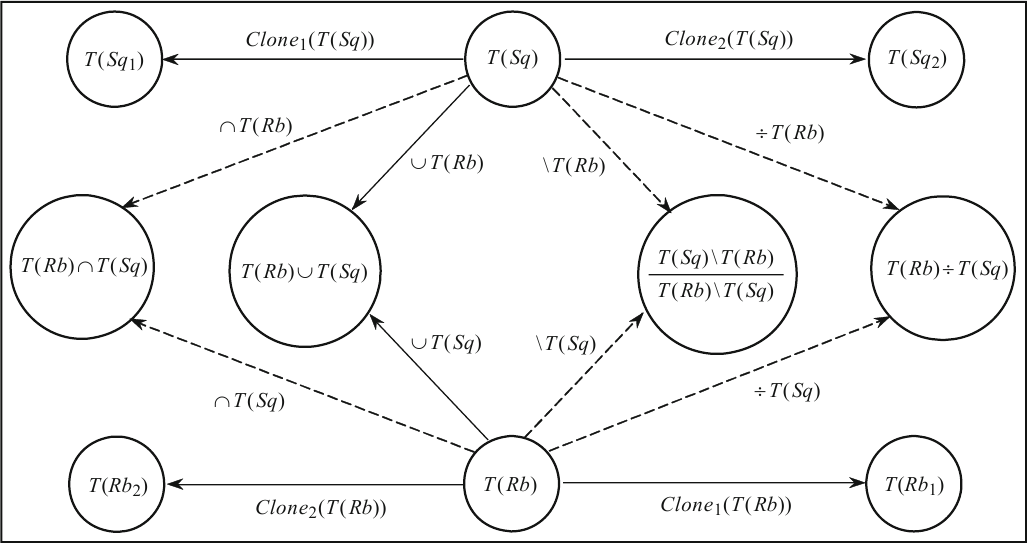}
\caption{Graphic representation of the principles of operation of exploiters for sets of objects and classes of objects.}
\label{fig-2}
\end{figure}

In Fig.~\ref{fig-1}, a fragment of a fuzzy object-oriented dynamic network without exploiters is shown. The principles of operation of the exploiters of the constructed FOODN are presented in Fig.~\ref{fig-2}.

In Fig.~\ref{fig-2}, a graph is presented that is a part of the FOODN shown in Fig.~\ref{fig-1}. The vertices of the former graph are classes of fuzzy objects. The classes $T(Rb)$ and $T(Sq)$ are of major importance since they are arguments for five types of exploiters from the set of exploiters $E$. All the other classes of objects are the results of using the classes $T(Rb)$ and $T(Sq)$ by the exploiters $E_1^n$, $E_2^n$, $E_3^2$, $E_4^2$, and $E_5^1$. To represent the result of each exploitation, several edges are used whose number equals the arity of the corresponding exploiter. For example, to represent the result of exploitation of the classes of objects $T(Rb)$ and $T(Sq)$ by the exploiter $E_1^2$, two edges $\cup(Rb)$ and $\cup(Sq)$ are used. A part of edges of the graph is marked by dashed lines in order to emphasize the fact that the intersection, difference, and symmetric difference of objects or classes of objects do not always exist \cite{Terletskyi-3}. Note that Fig.~\ref{fig-2} illustrates the results of using only two classes of fuzzy objects from the set $C$ by exploiters from the set $E$. Similar graphs can be constructed for any fuzzy objects from the set $O$ and classes of fuzzy objects from the set $C$. Exploiters can also be applied to fuzzy objects or classes of fuzzy objects obtained as a result of the previous use of exploiters with a view to generating new objects and classes of objects.

Modifiers and exploiters form the dynamic component of a network since the network can be extended and modified with their help and, as a result, new knowledge can be obtained and also its changes in time can be modeled.

In the presented example, only universal operations over objects and classes of objects were considered as exploiters. Note that the set of exploiters $E$, as well as the set of modifiers $M$, can consist of any methods that satisfy the corresponding definitions of an exploiter and a modifier that are presented in~\cite{Terletskyi-1}.

Exploiters are of great importance in object-oriented dynamic networks since their use allows one to create new objects and classes of objects of a network on the basis of objects and classes of objects underlying the network. Thus, exploiters allow one to obtain knowledge that is unobvious and thereby to expand the description of some object domain or other. The results of possible applications of exploiters to objects or classes of objects are given in Table~\ref{tab-5}.
\begin{table}
\centering{
\caption{Results of Different Applications of the Exploiters $E_1^n$, $E_2^n$, $E_3^2$, $E_4^2$, and $E_5^1$.}
\label{tab-5}
\begin{tabular}{p{1.5cm}p{5.2cm}p{5.2cm}}
\hline\noalign{\smallskip}
\multicolumn{1}{c}{Exploiter} & Results of Application to Objects & Results of Application to Classes of Objects\\
\noalign{\smallskip}
\hline
\noalign{\smallskip}
\multicolumn{1}{c}{$\cup$} & \multicolumn{1}{c}{A set of objects and class of objects} & \\  \cline{1-2}
\multicolumn{1}{c}{$\cap$} & &\\
\multicolumn{1}{c}{$\setminus$} & \multicolumn{1}{c}{A class of object} & \multicolumn{1}{c}{A class of objects}\\
\multicolumn{1}{c}{$\div$} & & \\\cline{1-2}
\multicolumn{1}{c}{$Clone_i$} & \multicolumn{1}{c}{An object} &\\
\hline
\end{tabular}}
\end{table}

Analyzing Table~\ref{tab-5}, one can see that the operations of intersection, difference, and symmetric difference make it possible to generate new classes of objects, and the cloning operation makes it possible to generate copies of objects or classes of objects. Of particular interest is the union operation that makes it possible to create sets of objects and classes of objects from elements of these sets. This approach to the creation of classes of objects is materially extensional, but, at the same time, it allows one to create heterogeneous classes, i.e., classes describing objects of different types. Such classes are immediately relevant to sets of objects that can consist of not only one-type elements. In this case, the class of objects that forms a set of objects is some prototype despite its extensional nature since, after defining such a class, new objects of this type can be constructed. Note that, within its framework, only the objects can be constructed that are equivalent to the objects belonging to the set of objects underlying this class. In many existing KRMs, the main representative components are objects, classes, and concepts that are represented by them. If sets of objects are used for knowledge representation, then, in contrast to objects, classes, and even concepts, they allow one to simultaneously consider some number of not necessarily one-type objects, which allows one to describe more complicated knowledge structures.

\section*{Conclusions}

This work presents the conception of the object-oriented approach to the representation of fuzzy knowledge. The generalization of object-oriented dynamic networks to the fuzzy case is proposed. The representation of fuzzy knowledge is illustrated by an example of construction of a fuzzy object-oriented dynamic network for some classes of fuzzy convex polygons. The described approach allows one to represent fuzzy knowledge and to model its changes in time and also provides a mechanism for obtaining new knowledge from basic knowledge; this mechanism considerably differs from well-known methods of knowledge acquisition in existing KRMs.

%
%

%
%

\end{document}